\documentclass[10pt,twocolumn,letterpaper]{article}

\usepackage{cvpr}              

\usepackage{mwe}
\usepackage{amsfonts}
\usepackage{sidecap}
\sidecaptionvpos{figure}{t}

\usepackage{booktabs}
\usepackage{siunitx}
\usepackage{makecell}

\newcommand{\myref}[2]{%
  \hyperref[#2]{#1~\ref*{#2}}%
}

\definecolor{hrcolor-links}{RGB}{193,38,1}
\definecolor{hrcolor-urls}{RGB}{9,46,171}
\definecolor{hrcolor-cite}{RGB}{47,143,0}

\usepackage{hyperref}
\hypersetup{
	breaklinks=true,
	pdfpagemode={UseOutlines}, 
	plainpages=false, 
	pdftitle={Homographic Navigation: Geometry-Driven Camera Guidance for Deterministic Planar Capture}, 
	pdfauthor={anonymized}, 
	pdfkeywords={}, 
	linkcolor=hrcolor-links, 
	citecolor=hrcolor-cite, 
	urlcolor=hrcolor-urls,
	colorlinks=true
}

\graphicspath{{images}}

\def\paperID{*****} 
\def\confName{CVPR}
\def\confYear{2026}

\title{Homographic Navigation:\\Geometry-Driven Camera Guidance for Deterministic Planar Capture}

\author{Dominik Kroupa$^{\dagger}$ \quad Marek Vaško$^{\dagger}$ \quad Muh Yuzril Ihza Baharuddin$^{\dagger\ast}$ \quad Adam Herout$^\dagger$
\vspace{0.5em}\\
$^\dagger$\,GRAPH@FIT, Brno University of Technology, Faculty of Information Technology \\
$^\ast$\,Ondokuz Mayıs University \\
\tt\small \{ikroupa, ivasko, herout\}@fit.vut.cz \\ \tt\small myuzrilib@gmail.com
}

\begin{document}
\maketitle

\begin{abstract}
We present \textbf{homographic navigation}, a geometry-centric framework for guiding camera acquisition toward precise capture of planar regions. Rather than treating homography as an output, we use it as an organizing variable that unifies learning, alignment, and evaluation. From a single annotated reference image, we generate unlimited synthetic training data via homographic augmentation and train a single-shot model for joint recognition and localization of multiple artifacts (physical objects with a rectangular planar target) through sparse keypoint prediction.

To address precision under limited model input resolution, we introduce a two-pass inference scheme with global detection followed by localized refinement, and a Stable Warp training strategy that significantly improves accuracy, particularly in the high-precision regime. The model also predicts confidence estimates per predicted keypoint and per the whole sample.

Experimental results demonstrate that accurate planar alignment can be achieved from minimal supervision, providing a foundation for geometry-driven camera guidance and future learning from in-the-wild video data.
\end{abstract}

\section{Introduction}
\label{sec:intro}

Capturing images of planar surfaces in a precise and repeatable manner is a fundamental requirement in many real-world applications, including condition monitoring, surface inspection, and artifact identification. In such settings, the goal is not merely to recognize or detect an object, but to acquire images under consistent geometric configurations so that meaningful comparisons can be made across time, devices, or environments. Achieving this level of consistency with a handheld camera, however, remains a challenging problem due to variations in viewpoint, scale, lighting, and partial or full occlusion (see~\myref{Figure}{fig:motivation}).

\begin{figure*}
    \centering
    \includegraphics[width=\linewidth]{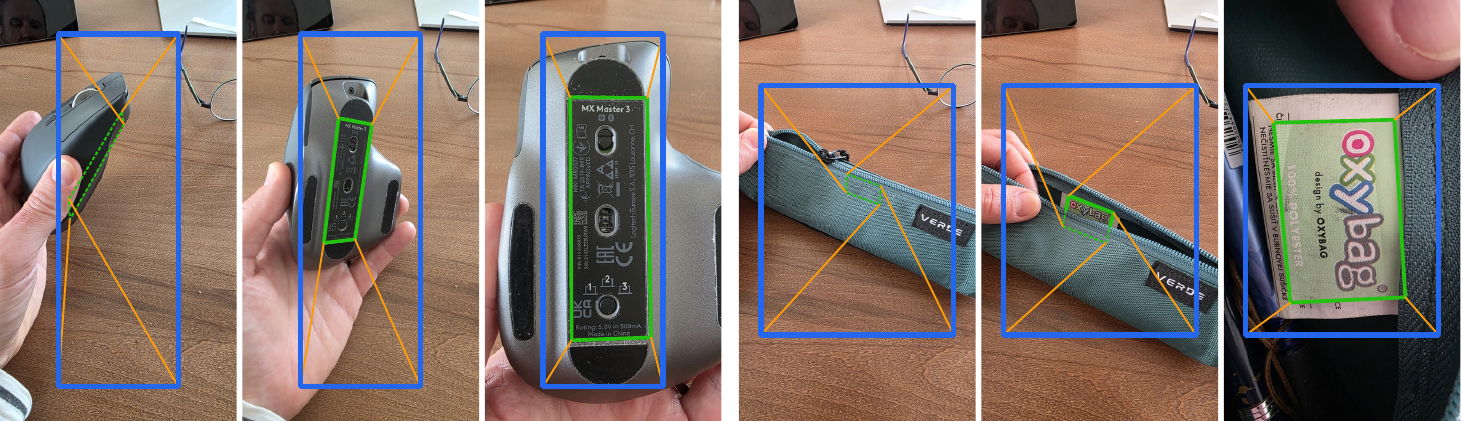}
    \caption{Homographic navigation should navigate the user to capture a planar surface even in challenging conditions -- including in-the-wild lighting and scene arrangement, and occlusion.}
    \label{fig:motivation}
\end{figure*}

A large body of prior work addresses related challenges through planar homography estimation. Classical approaches rely on feature detection and matching followed by robust estimation procedures, while more recent methods employ deep learning to directly regress homographies or learn intermediate correspondences. Recent advances include transformer-based architectures and iterative refinement models that significantly improve accuracy and robustness under challenging conditions such as large viewpoint changes and low-texture scenes~\cite{detone2016deep, zhu2024mcnet, he2026homofm, zhou2023transformer}. 

At the same time, planar alignment has been extensively studied in the context of planar object tracking and augmented reality. Despite substantial progress, recent benchmarks highlight that state-of-the-art methods still degrade significantly in unconstrained real-world conditions, particularly under occlusion, extreme viewpoint changes, and appearance variations~\cite{liu2023planartrack}. A comprehensive review further emphasizes that, despite decades of research, homography estimation remains sensitive to real-world variability and domain shifts~\cite{luo2023homography_review}.
Despite these advances, homography is typically treated as the \emph{output} of an inference process. While this enables alignment between images, it does not address \textbf{how an image should be acquired so that} a desired alignment is obtained reliably. In practice, this leads to unstable or inconsistent capture, especially when images are taken in unconstrained environments without explicit guidance.

In this work, we introduce \textbf{homographic navigation}, a geometry-centric framework in which homography is elevated from a prediction target to an organizing variable that jointly structures acquisition, learning, and evaluation. Rather than treating alignment as a passive estimation task, we consider how camera motion can be guided so that a planar region of interest is captured in a predefined geometric configuration. The central objective is therefore to enable a form of acquisition in which correct alignment becomes a direct consequence of the camera pose, rather than an uncertain post hoc estimate (\myref{Section}{sec:framework}).

Each physical object, \textbf{artifact}, defines a planar region of interest and is modeled at the instance level. The overall framework is organized as a two-phase process. In the first phase, a model is learned from a single reference image using synthetic homographic augmentation, enabling initial recognition and localization with minimal user input. In the second phase, the model is further improved by harvesting additional training data from unlabeled in-the-wild videos, where confident predictions are used to collect diverse real examples. This progressively expands the training distribution and supports curriculum-style learning, allowing the model to adapt to increasingly challenging conditions while maintaining geometric consistency.

\textbf{This paper focuses on the first phase} of the proposed framework, in which an artifact model is learned from a single reference image using geometry-driven augmentation. We employ a single-shot convolutional architecture based on YOLO~\cite{redmon2016yolo,tekin2018seamless} that predicts sparse keypoint configurations, together with objectness and confidence measures, enabling robust homography estimation from the predicted correspondences.

We evaluate the proposed approach using a homography-based protocol that measures reprojection error on independently annotated keypoints, ensuring that performance reflects true geometric alignment rather than direct regression accuracy on training targets. The results demonstrate that accurate planar localization can be achieved from extremely limited supervision.

In summary, the contributions of this work are:
\begin{itemize}
    \item We introduce homographic navigation, a framework that treats homography as an organizing variable for camera guidance, learning, and evaluation.
    \item We design a synthetic training pipeline (the first phase) based on homographic augmentation that supports efficient learning without additional labeling.
    \item We introduce a dataset spanning real, augmented real, and synthetic imagery with consistent homography annotations, supporting systematic evaluation of geometric accuracy, domain shift, and scale variability.
\end{itemize}

These results establish the feasibility of geometry-driven, minimally supervised learning for precise planar alignment, and form a foundation for future work on navigation-guided image acquisition.

\section{Related Work}

\subsection{Planar Alignment and Image-Based Augmented Reality}
Planar homography has long served as a principled abstraction for recovering the geometric relationship between a camera and a planar surface, enabling applications such as augmented reality and visual tracking. Early systems relied on engineered fiducial markers that provide reliable correspondences and stable pose estimation even under occlusion~\cite{garrido2014aruco}. Subsequent work shifted toward markerless approaches based on natural image targets, combining local feature detection, matching, and robust estimation to achieve real-time planar tracking in unconstrained scenes~\cite{wagner2010realtime}.

Despite their practical success, such methods remain sensitive to viewpoint changes, appearance variation, and partial occlusion, as confirmed by recent large-scale evaluations such as PlanarTrack~\cite{liu2023planartrack}. These limitations motivate approaches that move beyond purely passive alignment and instead consider how acquisition conditions themselves influence alignment quality.

\subsection{Learning-Based Homography Estimation}
Recent work has increasingly adopted deep learning to estimate planar homographies directly from image pairs. Early approaches demonstrated that convolutional networks can regress homography parameters end-to-end without explicit feature matching~\cite{detone2016deep}, while later methods improved robustness through unsupervised objectives and learned feature representations~\cite{nguyen2018unsupervised, zhang2020content}. 

A common challenge in this line of work is the reliance on synthetic training data. Recent approaches address the realism gap by generating more faithful training samples from real images, for example, by refining homography-based augmentation pipelines or iteratively improving training data quality~\cite{jiang2023supervised}. In parallel, iterative and recurrent architectures have been proposed to progressively refine homography estimates, improving accuracy under large transformations and difficult conditions~\cite{cao2022iterative, cao2023recurrent}. 

More recent models further incorporate multiscale reasoning, attention mechanisms, and efficient correlation structures to improve performance and efficiency~\cite{zhou2023transformer, zhu2024mcnet}. Very recent formulations reinterpret homography estimation as a continuous transformation process, improving robustness and domain generalization~\cite{he2026homofm}. 

Across these approaches, a common design pattern is that homography is treated as the \emph{output} of the model, either regressed directly or inferred from learned correspondences. In contrast, the present work treats homography as an organizing variable that structures training data generation, prediction, and evaluation.

\subsection{Planar Object Localization and Tracking}
Closely related problems include planar object detection, tracking, and alignment, where the goal is to localize and geometrically align planar regions such as documents, screens, or product surfaces. Recent approaches incorporate joint modeling of geometry, visibility, and confidence to improve robustness under occlusion and appearance changes~\cite{zhang2022hvcnet}. 

Nevertheless, these methods typically operate in a single-frame or short-term tracking regime, focusing on estimating alignment from arbitrary viewpoints. As demonstrated by recent benchmarks~\cite{liu2023planartrack}, performance degrades substantially under real-world variability, suggesting that inference-only formulations may be insufficient to guarantee reliable alignment. This motivates approaches that explicitly incorporate acquisition into the problem formulation.

\subsection{Camera Pose Estimation and Geometry-Aware Learning}
More generally, the problem of estimating camera pose from images has been widely studied in visual localization. Learning-based approaches such as PoseNet~\cite{kendall2015posenet} cast pose estimation as direct regression, but subsequent analyses show that such methods often struggle with accuracy and generalization compared to geometry-based approaches~\cite{sattler2019understanding}. 

These observations highlight a broader limitation of inference-centric formulations: predicting geometry from arbitrary inputs is fundamentally more challenging than constraining inputs to satisfy desired geometric conditions. Our approach follows this perspective and focuses on guiding acquisition so that correct alignment becomes a consequence of camera pose.

\subsection{Instance-Level Learning and Geometry-Driven Supervision}
Learning from limited supervision has been extensively explored in few-shot and instance-level learning, where strong inductive biases and prior structure are essential for effective generalization~\cite{song2023fewshot}. In parallel, self-supervised approaches exploit geometric constraints, such as homography consistency across views, to provide supervision without manual labels~\cite{muller2023self}. Confidence-based self-training strategies further enable models to expand training data by selectively incorporating reliable predictions while mitigating drift~\cite{cascante2021curriculum}. 

Test-time augmentation and prediction consistency have also been used as practical proxies for predictive uncertainty and robustness~\cite{conde2023tta}. The approach proposed in this work combines these ideas in a geometry-driven setting: planar constraints enable learning from a single reference image, while synthetic homographic augmentation and perturbation-based consistency provide both scalable supervision and a direct estimate of prediction reliability.

\section{Homographic Navigation Framework}
\label{sec:framework}

We introduce \emph{homographic navigation} as a framework for geometry-driven image acquisition, in which camera motion, learning, and alignment are unified through planar homography. The central idea is to move from inference-centric formulations, where alignment is estimated after capture, to acquisition-aware systems in which the camera is guided toward a configuration that directly yields the desired geometric alignment.

An overview of the framework is shown in \myref{Figure}{fig:homonav-framework}. The system is organized as a two-phase learning process centered on artifact-specific models. Each artifact is defined by a planar region of interest, specified by a single reference image and an axis-aligned bounding box -- AABB is used for simplicity of annotation; it could as well be a general quadrilateral. From this minimal input, the system learns to localize the region and estimate its geometric configuration under varying conditions.

\begin{figure}[t]
    \centering
    \includegraphics[width=\linewidth]{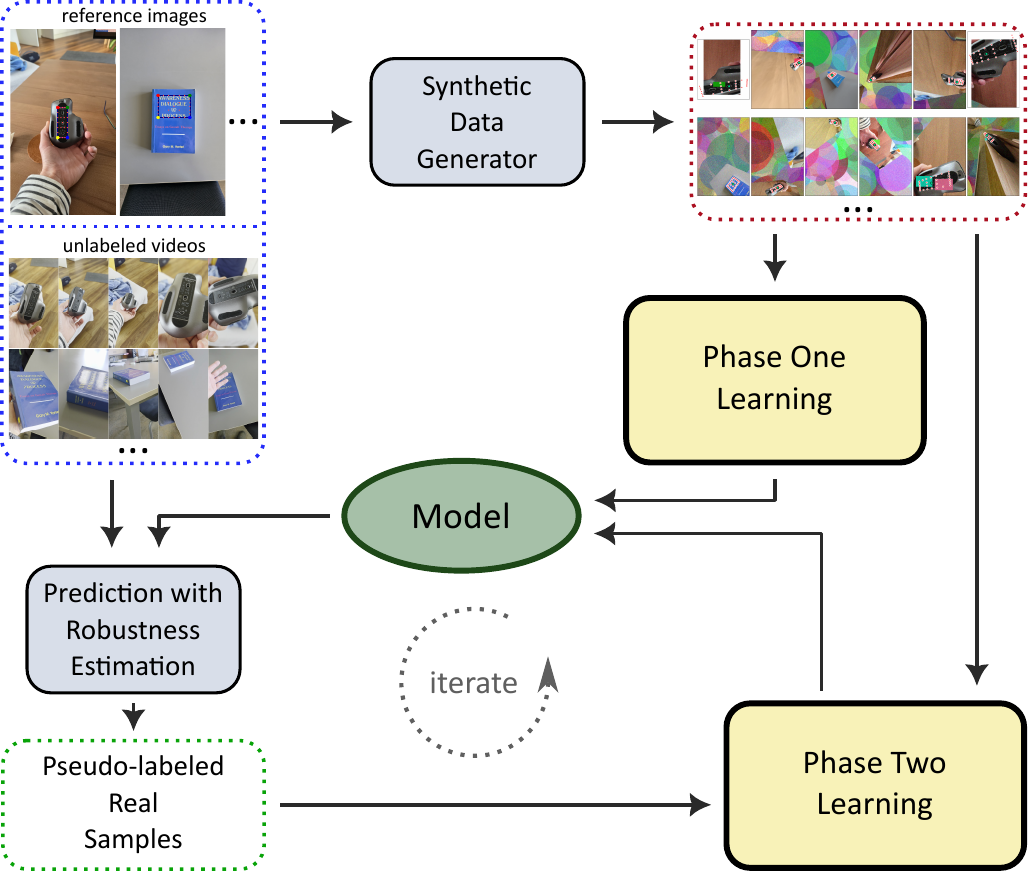}
    \caption{Overview of the homographic navigation framework. Phase One (this paper) learns an artifact-specific model from a single reference image using synthetic homographic augmentation. In later phases, the model is extended using pseudo-labeled samples from unlabeled videos, enabling iterative self-supervised learning and eventual camera guidance toward a desired geometric configuration.}
    \label{fig:homonav-framework}
\end{figure}

The present work addresses the first phase of this framework, referred to as \emph{Phase One}. In this phase, a model is learned from a single reference image using geometry-driven synthetic data generation. Homographic transformations are used to produce a wide range of training samples that simulate viewpoint changes, scale variation, and partial visibility, enabling reliable planar localization without additional manual annotation.

Homography plays a central role as the organizing variable of the framework. It defines the relationship between reference and input views, provides supervision through synthetic data generation, and serves as the evaluation metric via geometric alignment. In later stages, it also becomes the target of camera guidance, enabling navigation toward a desired capture configuration.

The learned model predicts sparse keypoints and associated confidence measures, allowing robust estimation of the homography between images. This establishes the representation and reliability estimates required for subsequent stages, including self-supervised learning from video and closed-loop camera navigation.

\section{Method: Phase One -- Learning from a Single Image via Homographic Augmentation}

\subsection{Synthetic Training via Homographic Augmentation}

\begin{figure*}
    \centering
    \includegraphics[height=55mm]{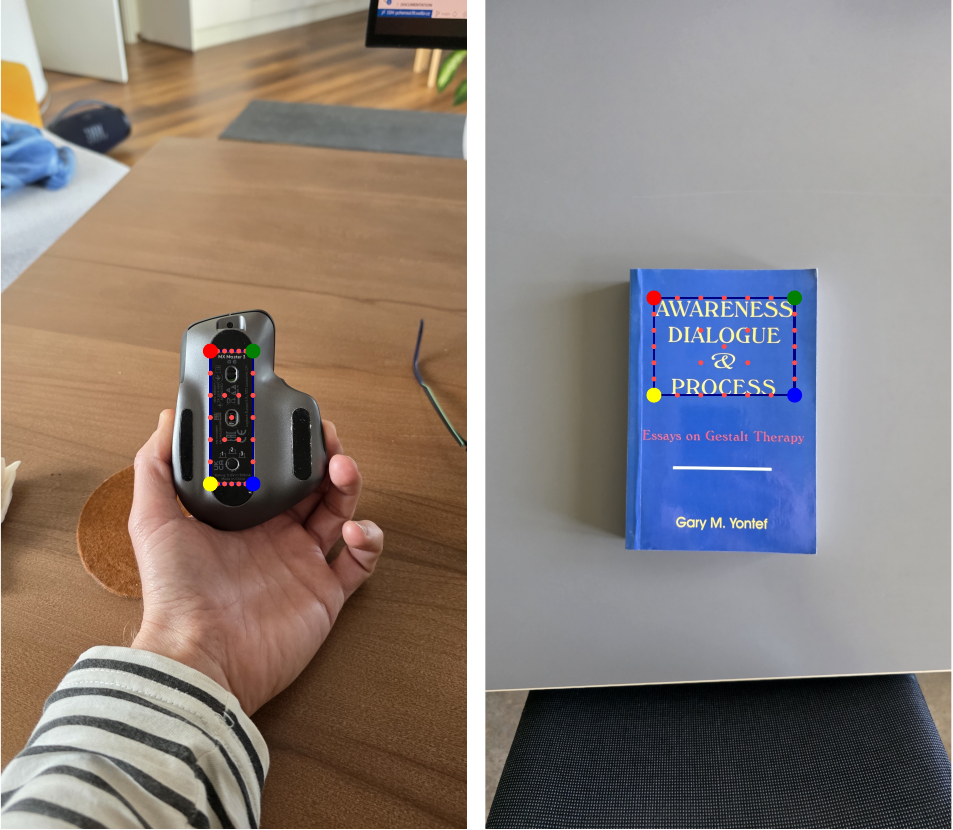}\hfill
    \includegraphics[height=55mm]{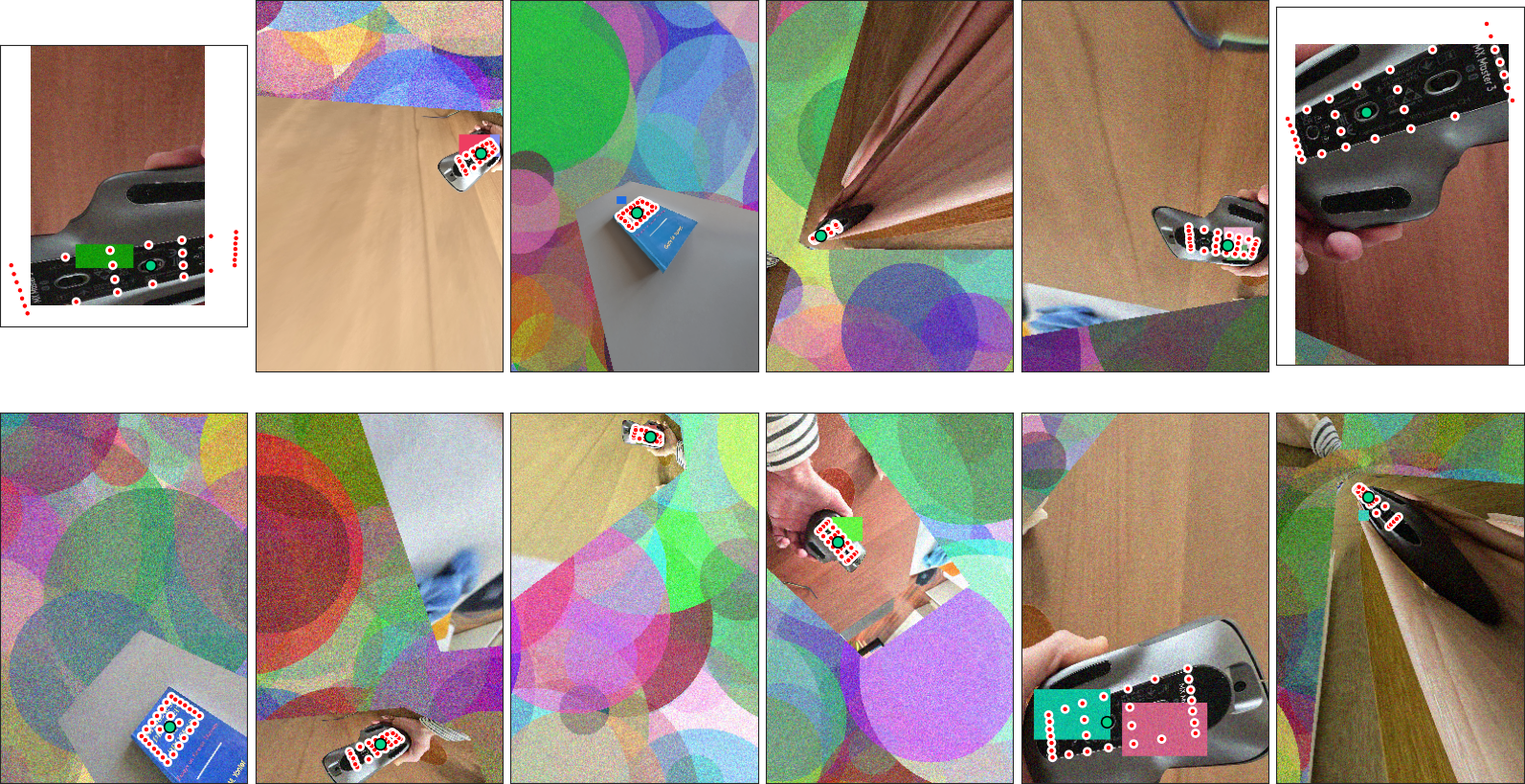}
    \caption{Two examples of artifacts and the generated training samples. \textbf{left:} Reference images with the defined axis-aligned bounding box and the generated keypoints. \textbf{right:} Examples of generated training samples.  Note that the keypoints can extend beyond the training images.}
    \label{fig:artifacts}
\end{figure*}

Given a reference image $I_s$ of an artifact (see \myref{Figure}{fig:artifacts} left for two examples) and a planar region of interest user-defined by an axis-aligned rectangular bounding box
\begin{equation}
\mathbf{c} = \{(x_L, y_T), (x_R, y_T), (x_R, y_B), (x_L, y_B)\} \subset \mathbb{R}^2,
\end{equation}
we construct a training process based on synthetic view generation under planar projective transformations.

For each training sample, a target quadrilateral $\mathbf{d} = \{\mathbf{d}_i\}_{i=1}^{4}$ is generated by applying a sequence of stochastic transformations to the initial axis-aligned rectangle $\mathbf{c}$. These transformations include random perturbations of corner positions, global rotation, scaling, and translation, allowing controlled variation in viewpoint, orientation, apparent size, and partial visibility. A homography $\mathbf{H} \in \mathbb{R}^{3 \times 3}$ is then computed such that
\begin{equation}
\tilde{\mathbf{d}}_i = \mathbf{H}\, \tilde{\mathbf{c}}_i, \quad i = 1, \dots, 4,
\end{equation}
where $\tilde{(\cdot)}$ denotes homogeneous coordinates and $\lambda$ is a scale factor. The target image of a resolution matching the CNN model is generated by warping the reference image using the inverse mapping
$I_t(\mathbf{x}) = I_s\!\left(\mathbf{H}^{-1} \mathbf{x}\right)$,
with appropriate interpolation and background compositing.

In addition to the corner points, we define a fixed set of canonical keypoints $\mathbf{K} \in \mathbb{R}^{K \times 2}$ within the rectangular region. These keypoints are deterministically derived from the bounding box geometry (they happen to include the four corner points) and provide a sparse representation of the planar structure; $K=29$ in the current implementation and in the experiments. For each synthetic sample, the corresponding target keypoints are obtained by applying the same homography:
$\mathbf{K}' = \pi\!\left(\mathbf{H}\, \tilde{\mathbf{K}}\right)$,
where $\pi(\cdot)$ denotes dehomogenization. The transformed keypoints serve as supervised targets for learning.

Additional stochastic variations are further applied after compositing and warping to increase appearance diversity while preserving geometric consistency. Specifically, the synthesized image is subjected to mild photometric perturbations, including contrast and brightness adjustments and additive Gaussian noise with randomly varying intensity. In addition, we apply structured spatial dropout, in which one or two contiguous rectangular regions are randomly masked within an extended bounding box defined around the projected keypoints. The masks are sampled with probability $p$ (and $p^2$ for a second block), with sizes drawn as fractions of the keypoint extent, and are filled with constant values. By conditioning the masking on the keypoint region, the augmentation encourages robustness to structured occlusion while maintaining relevance to the task.

By sampling a new homography and background for each sample in each iteration, this process defines an effectively unbounded distribution of training examples. This geometry-driven augmentation enables efficient training from a single annotated image while exposing the model to a broad range of viewing conditions relevant for planar alignment.

\subsection{Model Architecture}

The proposed model (\myref{Figure}{fig:cnn-architecture}) follows a single-shot convolutional architecture inspired by Tekin \etal~\cite{tekin2018seamless} that maps an input image $X \in \mathbb{R}^{H \times W \times 3}$ to a dense prediction grid of geometric and semantic outputs. In the current setting, the input resolution is $H \times W = 384 \times 256$, resulting in a grid of size $G_h \times G_w = 12 \times 8$. The network is composed of a backbone feature extractor, a multi-resolution neck, and a convolutional prediction head, and produces all outputs in a single forward pass.

A ResNet-18 backbone (up to layer 3) is used to extract intermediate feature maps at multiple spatial resolutions. In particular, we use a feature tensor $F \in \mathbb{R}^{24 \times 16 \times C}$ (i.e., resolution $H/16 \times W/16$) as input to the neck. The neck module combines a main processing branch with a parallel high-frequency preservation branch. The main branch applies spatial downsampling followed by a sequence of convolutional transformations,
$F_{\mathrm{main}} = \phi_{\mathrm{main}}(F)$,
producing a feature map at resolution $G_h \times G_w$ with high channel capacity. In parallel, a side branch applies a $1\times1$ convolution $\psi(\cdot)$ that compresses the channel dimension of $F$ and applies a pixel unshuffle operation, which rearranges local spatial neighborhoods into channel dimensions without information loss:
\begin{equation}
F_{\mathrm{side}} = \mathrm{unshuffle}\!\left(\psi(F)\right).
\end{equation}
The resulting tensors are concatenated along the channel dimension,
yielding a representation that jointly captures coarse semantic structure and fine spatial detail.

The prediction head consists of a small stack of $3\times3$ convolutional layers followed by a $1\times1$ projection to the final output tensor
\begin{equation}
Y \in \mathbb{R}^{G_h \times G_w \times (2 + 3K + C)},
\end{equation}
where $K$ is the number of keypoints and $C$ is the number of artifact classes ($C=30$ in the experiments). For each grid cell, the output encodes an objectness score, a sample-level confidence, and for each keypoint a triplet $(\Delta x, \Delta y, q)$ corresponding to a cell-relative offset and confidence, together with class logits.

For comparison, a simpler \emph{straight neck} variant, consisting of a single processing branch without multi-resolution fusion and without pooling, is also implemented.  Since it produces spatially twice as large an output, it can achieve slightly better accuracy at the cost of considerably greater computational complexity (14.6 vs. 6.0 estimated GFLOPs at the same input resolution). The multi-resolution design is preferred as it preserves high-frequency spatial information while maintaining a compact representation.

\begin{figure*}
    \centering
    \includegraphics[width=0.9\linewidth]{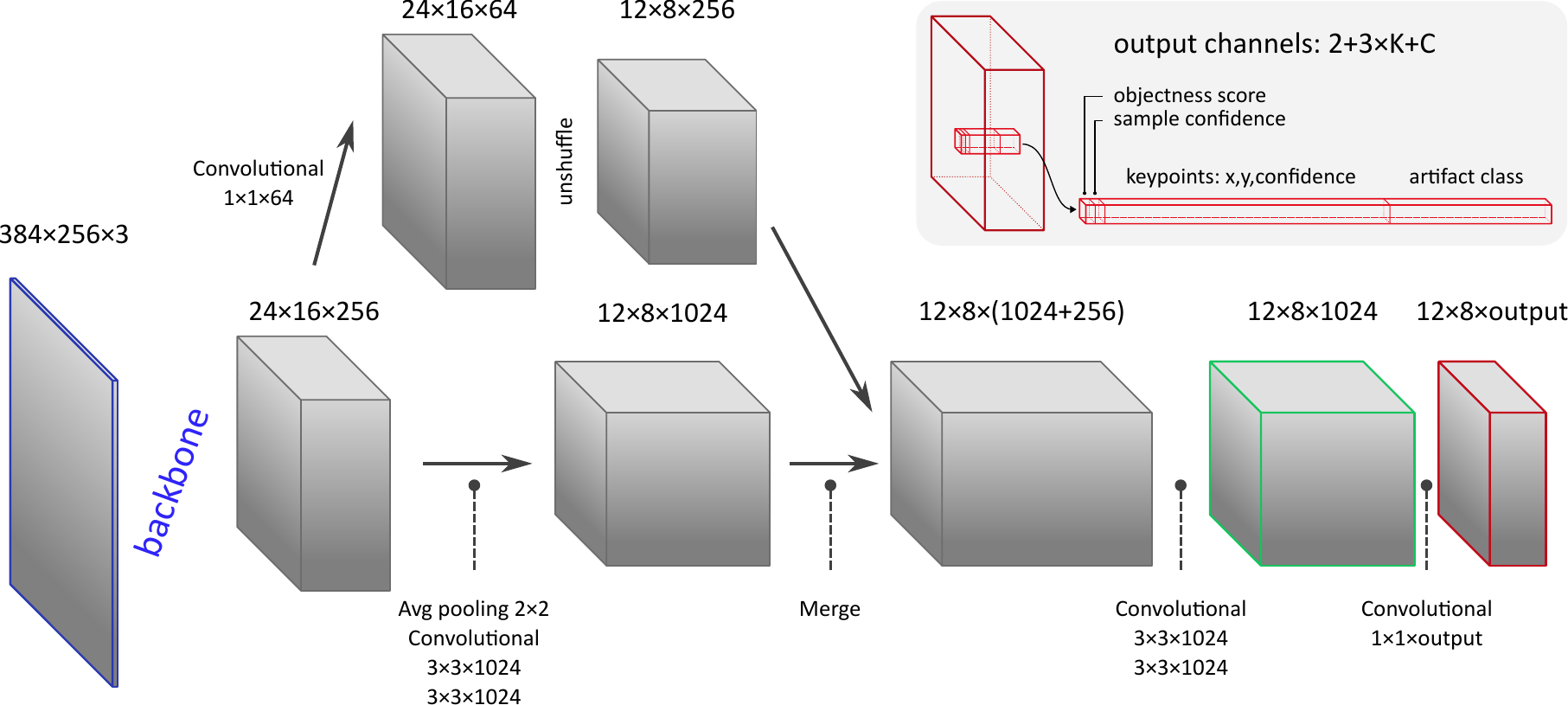}
    \caption{The backbone (ResNet-18) produces intermediate features at $24 \times 16$ resolution, which are processed by the multi-resolution neck consisting of a main branch and a parallel detail-preserving branch based on channel compression and pixel unshuffle. The fused representation is mapped to a dense $G_h \times G_w$ prediction grid encoding objectness, per-keypoint offsets and confidences, sample confidence, and class logits.}
    \label{fig:cnn-architecture}
\end{figure*}

\subsection{Loss Function}
\label{sec:loss}

The model is trained using a composite loss that supervises multiple complementary objectives on the output grid of size $G_h \times G_w$. Each grid cell predicts objectness, keypoint offsets, confidence estimates, and class logits. Let $\mathcal{S}$ denote the set of supervised cells associated with a target sample, defined by the cell containing the object centroid (\emph{responsible cell}) and its \emph{neighborhood} (4- or 8-connected). The total loss is given by
\begin{equation}
\mathcal{L} = \mathcal{L}_{\mathrm{obj}} 
+ \lambda_{\mathrm{off}} \mathcal{L}_{\mathrm{offset}} 
+ \mathcal{L}_{\mathrm{cls}} 
+ \lambda_{\mathrm{kp}} \mathcal{L}_{\mathrm{kp\text{-}conf}} 
+ \lambda_{\mathrm{s}} \mathcal{L}_{\mathrm{sample\text{-}conf}},
\end{equation}
where the weights $\lambda_{\mathrm{off}}, \lambda_{\mathrm{kp}}, \lambda_{\mathrm{s}}$ control the relative contributions of the individual components.

\paragraph{Objectness Learning}
A binary objectness map $o^* \in \{0,1\}^{G_h \times G_w}$ is defined such that $o^*_{uv}=1$ only at the responsible cell and $0$ elsewhere. Objectness is optimized using binary cross-entropy with per-cell weighting:
\begin{equation}
\mathcal{L}_{\mathrm{obj}} = 
\frac{\sum_{u,v} w_{uv} \cdot \mathrm{BCE}(\hat{o}_{uv}, o^*_{uv})}
     {\sum_{u,v} w_{uv}},
\end{equation}
where the responsible cell is assigned weight $1$, background cells are down-weighted, and neighboring cells are excluded from supervision.

\paragraph{Keypoint Regression}
For each supervised cell $(u,v)\in\mathcal{S}$ and each keypoint $k$, the model predicts offsets $(\hat{\Delta} x_k, \hat{\Delta} y_k)$ relative to the cell. These are regressed to ground-truth offsets using a Smooth-$\ell_1$ loss:
\begin{equation}
\mathcal{L}_{\mathrm{offset}} =
\frac{1}{|\mathcal{S}|K}\sum_{(u,v)\in\mathcal{S}} \sum_{k=1}^{K}
\mathrm{SmoothL1}\!\left(
\hat{\Delta}^{(u,v)}_{k},\; \Delta^{*(u,v)}_{k}
\right).
\end{equation}

\paragraph{Confidence Estimation}
The model predicts both per-keypoint confidences $\hat{q}_k^{(u,v)}$ and a per-cell sample confidence $\hat{s}^{(u,v)}$. Their targets are derived from the geometric prediction error. For keypoints:
\begin{equation}
q_k^{*(u,v)} = f\!\left(d_k^{(u,v)}\right),
\end{equation}
where $d_k^{(u,v)}$ is the Euclidean distance (in pixels) between the predicted and ground-truth position of keypoint $k$. The confidence target is defined by a distance-decaying function
\begin{equation}
    f(d) =
    \begin{cases}
    \dfrac{\exp\!\left(\alpha \left(1 - \dfrac{d}{d_{\mathrm{th}}}\right)\right) - 1}{\exp(\alpha) - 1} & \text{if } d < d_{\mathrm{th}}, \\[6pt]
    0 & \text{otherwise,}
    \end{cases}
    \end{equation}
where $d_{\mathrm{th}}$ is a distance threshold and $\alpha$ controls the slope of the decay (in our experiments, $d_{\mathrm{th}} = 5$ pixels and $\alpha = 1$). This assigns high confidence to accurate predictions and suppresses confidence beyond the threshold.

The per-keypoint confidence loss is defined as
\begin{equation}
\mathcal{L}_{\mathrm{kp\text{-}conf}} =
\frac{1}{|\mathcal{S}|K} \sum_{(u,v)\in\mathcal{S}} \sum_{k=1}^{K}
\mathrm{SmoothL1}(\hat{q}_k^{(u,v)}, q_k^{*(u,v)}).
\end{equation}

Similarly, the sample-level confidence target is computed from the RMS keypoint error
\begin{equation}
d_{\mathrm{rms}}^{(u,v)} =
\sqrt{\frac{1}{K} \sum_{k=1}^{K} \left(d_k^{(u,v)}\right)^2},
\end{equation}
it is mapped through the same function
\begin{equation}
s^{*(u,v)} = f\!\left(d_{\mathrm{rms}}^{(u,v)}\right)
\end{equation}
and optimized as
\begin{equation}
\mathcal{L}_{\mathrm{sample\text{-}conf}} =
\frac{1}{|\mathcal{S}|}\sum_{(u,v)\in\mathcal{S}}
\mathrm{SmoothL1}(\hat{s}^{(u,v)}, s^{*(u,v)}).
\end{equation}
Gradients from confidence losses do not propagate into the offset predictions.

\paragraph{Classification Loss}
For each supervised cell $(u,v)\in\mathcal{S}$, the model predicts class logits $\hat{\mathbf{l}}^{(u,v)} \in \mathbb{R}^C$. Given the ground-truth class label $c^*$, the classification loss is:
\begin{equation}
\mathcal{L}_{\mathrm{cls}} =
\frac{1}{|\mathcal{S}|} \sum_{(u,v)\in\mathcal{S}}
\mathrm{CE}\left(\hat{\mathbf{l}}^{(u,v)}, c^*\right).
\end{equation}

\begin{figure*}[t]
    \centering
    \includegraphics[width=0.165\textwidth]{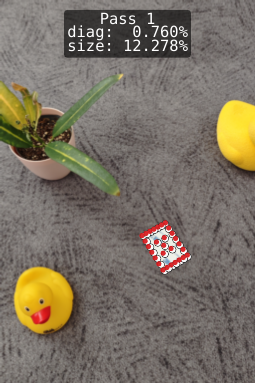}%
    \includegraphics[width=0.165\textwidth]{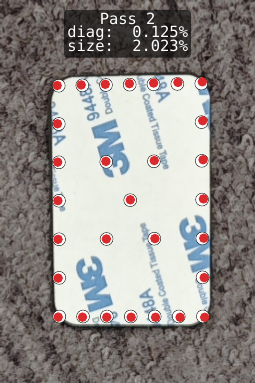}\hfill
    \includegraphics[width=0.165\textwidth]{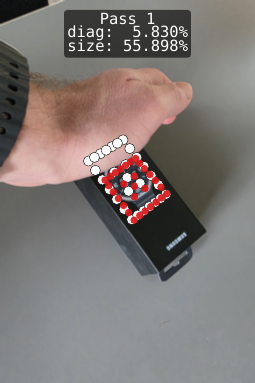}%
    \includegraphics[width=0.165\textwidth]{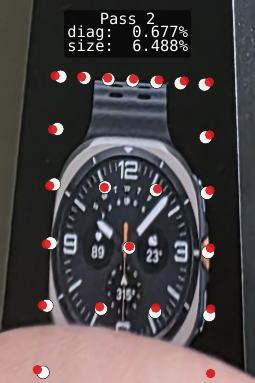}\hfill
    \includegraphics[width=0.165\textwidth]{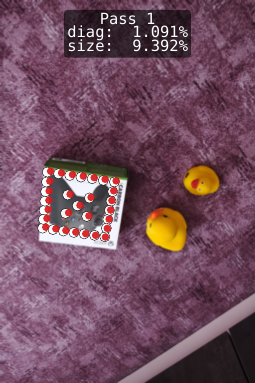}%
    \includegraphics[width=0.165\textwidth]{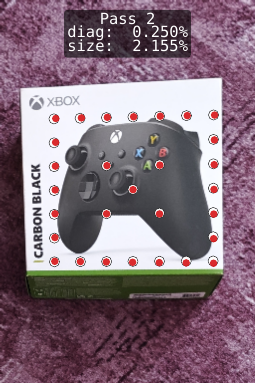}
    \caption{Two-pass inference. Pass~1 localizes coarsely; pass~2 refines on the target region. Red/white markers denote predicted/ground-truth keypoints. \emph{diag} and \emph{size} are normalized errors.}
    \label{fig:two-passes}
\end{figure*}

\subsection{Inference in Two Passes and Stable Warp Training}
\label{sec:two-pass-inference}

To preserve geometric precision under a fixed input resolution, inference is performed in two passes (\myref{Figure}{fig:two-passes}). A global pass processes a resized full image $I_1=\mathcal{P}_1(I)$ and yields coarse predictions
$(\hat{\mathbf{K}}_1,\hat{c}_1)=f_\theta(I_1)$.
These are used to define a focus transform that re-samples directly from the original high-resolution image,
\begin{equation}
I_2=\mathcal{P}_2(I;\hat{\mathbf{K}}_1,\mathbf{K}^{\star},m).
\end{equation}
Here, $\mathbf{K}^{\star}=\{\mathbf{k}_i^{\star}\}_{i=1}^N$ is a fixed set of canonical keypoints defined in a class-specific normalized coordinate frame (centered, upright, and at a standard scale), with each index corresponding to a consistent semantic location and $m$ is a margin parameter controlling the spatial extent of the focus region, i.e., how much the region inferred from $\hat{\mathbf{K}}_1$ is expanded before warping. The transform $\mathcal{P}_2$ maps the coarse prediction $\hat{\mathbf{K}}_1$ toward this canonical layout.

A second forward pass produces refined keypoints $(\hat{\mathbf{K}}_2,\hat{c}_2)=f_\theta(I_2)$, and final geometry is estimated as $\hat{\mathbf{H}}=\mathrm{Homography}(\hat{\mathbf{K}}_2,\mathbf{K}^{\star})$. This decomposition separates coarse discovery from fine localization, while avoiding loss of detail by re-sampling from $I$.

In the evaluation below, both passes are always executed and reported. In practical deployment (e.g., mobile AR), the system operates in a tracking regime: once the target is detected by the first pass, subsequent frames are processed only via the focused second-pass transformation centered at the previously estimated location. This exploits temporal continuity, as camera motion is typically smooth. If confidence drops (target lost), the system falls back to the global first pass to reinitialize. Thus, the model is not executed twice per frame in steady-state operation.

To support both regimes with a single model, we introduce \emph{Stable Warp} training. Each training sample is generated either by a wide-range augmentation $\mathcal{A}_{\text{wild}}$ or by a near-canonical augmentation $\mathcal{A}_{\text{stable}}$, with probabilities $1-\rho$ and $\rho$, respectively (with $\rho\approx 0.3$). The former ensures robustness for global discovery (pass 1), while the latter concentrates training signal in the high-precision regime induced by focused re-sampling (pass 2). This alignment enables a single network to operate effectively both in full-image search and in localized tracking, improving keypoint accuracy and homography estimation without increasing practical inference cost.

\section{Experiments and Results}

\begin{table*}[t]
    \centering
    \caption{Quantitative comparison of methods using AUC of cumulative error curves (higher is better). Results are reported for real, augmented real, and synthetic datasets under diagonal- and target-size-normalized error measures. FPS was measured on preloaded inputs using an AMD Ryzen 9 7900X CPU and an NVIDIA RTX 4080 16GB GPU.}
    \label{tab:results}
    \resizebox{\linewidth}{!}{
        \begin{tabular}{
            l
            !{\vrule} S S S
            !{\vrule} S S S
            !{\vrule} S S S
        }
            \toprule
            \multicolumn{1}{c}{} 
             & \multicolumn{3}{c}{\textbf{real}} 
             & \multicolumn{3}{c}{\textbf{real augmented}} 
             & \multicolumn{3}{c}{\textbf{synthetic}} \\
            
            \cmidrule(lr){2-4} \cmidrule(lr){5-7} \cmidrule(lr){8-10}
            
            \multicolumn{1}{c}{}
             & \multicolumn{2}{c}{\textbf{AUC} $\uparrow$}
             & \multicolumn{1}{c}{}
             & \multicolumn{2}{c}{\textbf{AUC} $\uparrow$}
             & \multicolumn{1}{c}{}
             & \multicolumn{2}{c}{\textbf{AUC} $\uparrow$}
             & \multicolumn{1}{c}{} \\
             
            \cmidrule(lr){2-3} \cmidrule(lr){5-6} \cmidrule(lr){8-9}
            
            \multicolumn{1}{c}{\textbf{method}} 
             & \multicolumn{1}{c}{\textbf{diag}} 
             & \multicolumn{1}{c}{\textbf{size}} 
             & \multicolumn{1}{c}{\textbf{fps}}
             & \multicolumn{1}{c}{\textbf{diag}} 
             & \multicolumn{1}{c}{\textbf{size}} 
             & \multicolumn{1}{c}{\textbf{fps}}
             & \multicolumn{1}{c}{\textbf{diag}} 
             & \multicolumn{1}{c}{\textbf{size}} 
             & \multicolumn{1}{c}{\textbf{fps}} \\
                    \midrule
            baseline: SIFT
            & 42.13 & 34.41 & 35.46
            & 38.39 & 30.22 & 36.91
            & 23.17 & 8.70 & 60.06 \\
        
            baseline: RoMa v2
            & 35.68 & 27.22 & 2.58
            & 35.57 & 25.41 & 2.58
            & 15.77 & 6.26 & 2.57 \\
        
            HomoNav
            & 33.39 & 25.04 & 289.91
            & 30.66 & 20.73 & 332.20
            & 47.52 & 36.27 & 323.40 \\
        
        

            w/o Stable Warp
            & 34.04 & 23.30 & 289.91
            & 31.03 & 19.09 & 332.20
            & 46.70 & 31.08 & 323.40 \\

            Multi-Res Neck
            & 33.79 & 20.98 & 430.75
            & 30.94 & 16.17 & 475.85
            & 47.86 & 35.54 & 482.19 \\
        \bottomrule
        \end{tabular}
        }
\end{table*}

\subsection{Training Details}

The main model is trained in a multi-class setting for $300{,}000$ iterations with batch size $64$ using Adam. Different learning rates are used for parameter groups ($10^{-4}$ for the backbone, $10^{-3}$ for the remaining layers), with a short linear warmup followed by cosine annealing to zero.

Artifact-specific specialization is performed rather as an auxiliary experiment. Starting from the trained multi-class checkpoint, each class is fine-tuned independently for $500$ iterations using a single learning rate $10^{-5}$ and cosine annealing. No architectural or pipeline changes are introduced in this stage.

\subsection{Datasets: Real, Real Augmented, Synthetic}

We evaluate on three complementary datasets built over the same set of $23$ artifact classes, each defined by a reference image with annotated keypoints and a canonical coordinate frame.  All datasets, together with the evaluation protocol and code, will be made publicly available upon publication.

\textbf{Real.}
A small but challenging dataset of $247$ manually annotated smartphone images captured under unconstrained conditions (cluttered backgrounds, varying lighting, viewpoint, and scale). Images are high-resolution ($\sim\!9$\,MP), and the artifact may occupy only a small portion of the frame. Ground-truth homographies are estimated from sparse manual keypoint annotations. This dataset serves as the primary evaluation benchmark.

\textbf{Synthetic.}
A procedurally generated dataset of $2000$ images created by homographic warping of reference images onto random backgrounds, followed by photometric augmentation and simulated occlusion. It provides exact ground-truth geometry and broad coverage of viewpoint and scale variation. This dataset is used for controlled evaluation and ablation.

\textbf{Real Augmented.}
A dataset of $2000$ images derived from real captures by applying additional homographic perturbations while preserving photorealistic appearance. Keypoints are propagated analytically through the transformations, yielding consistent annotations. This dataset bridges the gap between real and synthetic domains and tests generalization under moderate geometric variation.

\begin{figure}
    \centering
    \includegraphics[width=\linewidth]{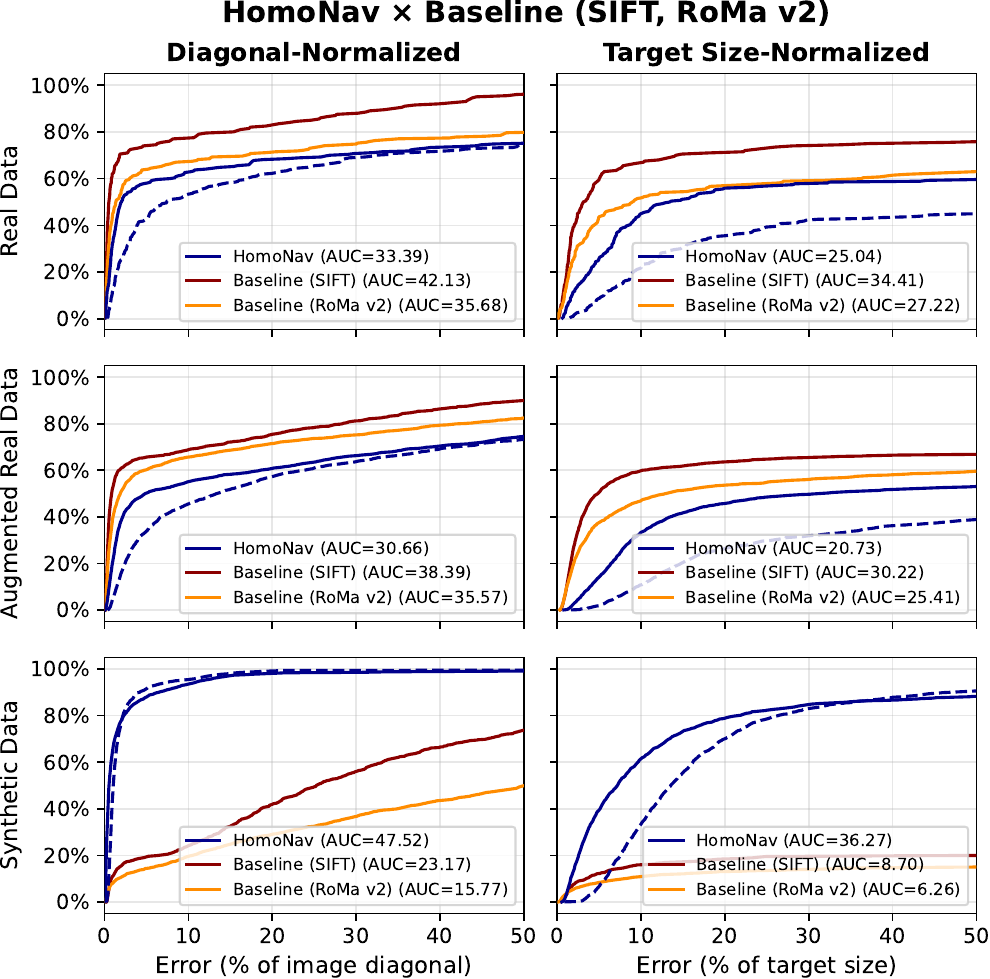}
    \caption{Comparison of HomoNav with SIFT+RANSAC and RoMa v2 using cumulative error curves across real, augmented, and synthetic datasets. Solid lines denote pass~2 and dashed lines pass~1 (\myref{Section}{sec:two-pass-inference}). Higher, more left-shifted curves indicate better performance; AUC values are summarized in \myref{Table}{tab:results}. Note that the baselines operate with known artifact identity and high-resolution templates, whereas HomoNav jointly recognizes and localizes all artifacts from downscaled inputs.}
    \label{fig:plot-baseline}
\end{figure}

\subsection{Evaluation Methodology}

All methods are evaluated using a common geometric protocol, ensuring direct comparability across HomoNav variants and baselines. Each sample provides an image $I$ and a ground-truth homography $\mathbf{H}^{\mathrm{gt}}_{I\rightarrow C}$ mapping image coordinates to the canonical frame defined by reference keypoints.

\paragraph{HomoNav evaluation.}
For each image, both inference passes (\myref{Section}{sec:two-pass-inference}) are executed:
\begin{equation}
(\hat{\mathbf{K}}_1,\hat c_1)=f_\theta(\mathcal P_1(I)),\quad
(\hat{\mathbf{K}}_2,\hat c_2)=f_\theta(\mathcal P_2(I;\hat{\mathbf{K}}_1,\mathbf{K}^\star,m)).
\end{equation}
Predicted keypoints are converted to homographies via robust estimation,
\begin{equation}
\hat{\mathbf H}^{(p)}_{I\rightarrow C}=\mathrm{Homography}(\hat{\mathbf K}_p,\mathbf K^\star),\quad p\in\{1,2\},
\end{equation}
and both pass-1 and pass-2 results are reported.

\paragraph{Baseline evaluation.}
Feature-based baselines (e.g.\ SIFT/RANSAC, RoMa~v2~\cite{edstedt2025romav2harderbetter}) first estimate a homography in a high-resolution rescaled image domain, which is mapped back to original resolution and composed with the known target-to-canonical transform to obtain $\hat{\mathbf H}_{I\rightarrow C}$. Failures to estimate a homography are treated as missing values.

\paragraph{Geometric error metric.}
Accuracy is measured by comparing canonical reprojections. Let $\{\mathbf q_j\}_{j=1}^4$ be four fixed canonical reference points (in the canonical space of $\mathbf{K}^*$). Using inverse homographies,
\begin{equation}
  \hat{\mathbf x}_j = \pi(\hat{\mathbf H}^{-1}_{I\rightarrow C}\tilde{\mathbf q}_j),\quad
  \mathbf x^{\mathrm{gt}}_j = \pi((\mathbf H^{\mathrm{gt}}_{I\rightarrow C})^{-1}\tilde{\mathbf q}_j),
\end{equation}
where $\pi(\cdot)$ denotes dehomogenization, we define a single RMSE-style aggregate pixel error
\begin{equation}
  E_{\mathrm{px}} = \sqrt{\sum_{j=1}^{4}\|\hat{\mathbf x}_j-\mathbf x^{\mathrm{gt}}_j\|_2^2}.
\end{equation}

\paragraph{Normalized error measures.}
To account for scale, we report two normalized errors:
\begin{equation}
  E_{\mathrm{diag}} = 100 \frac{E_{\mathrm{px}}}{\sqrt{W^2+H^2}},\qquad
  E_{\mathrm{size}} = 100 \frac{E_{\mathrm{px}}}{\sqrt{A_{\mathrm{gt}}}},
\end{equation}
where $(W,H)$ is image size and $A_{\mathrm{gt}}$ is the area of the ground-truth projected quadrilateral. The former reflects error relative to image extent, while the latter normalizes by object scale. Reporting both prevents bias toward either large or small targets.

\subsection{Comparison with Baseline: SIFT/RANSAC and RoMa v2}

\begin{figure}
    \centering
    \includegraphics[width=\linewidth]{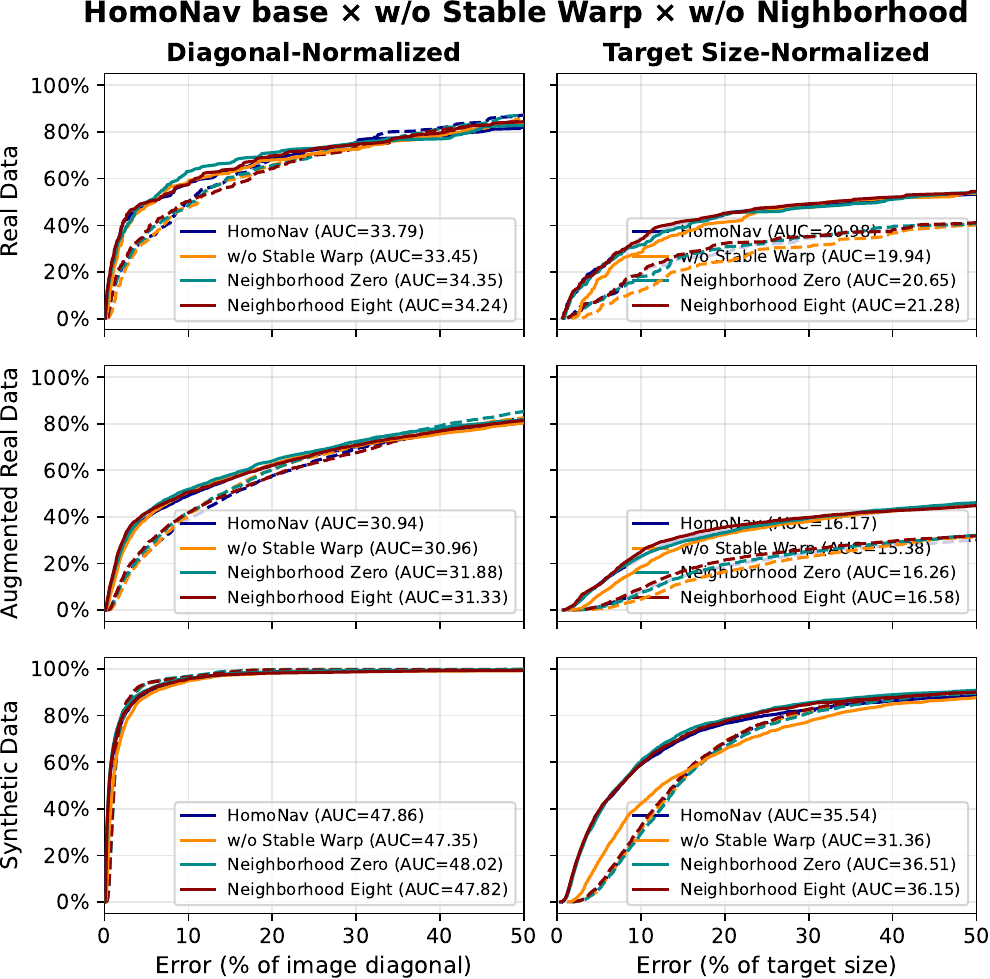}
    \caption{Comparison of training strategies using cumulative error curves. Influence of Stable Warp (\myref{Section}{sec:two-pass-inference}) and neighborhood supervision; Stable Warp notably improves pass~2 accuracy.}
    \label{fig:ablation-training}
\end{figure}

\begin{figure}
    \centering
    \includegraphics[width=\linewidth]{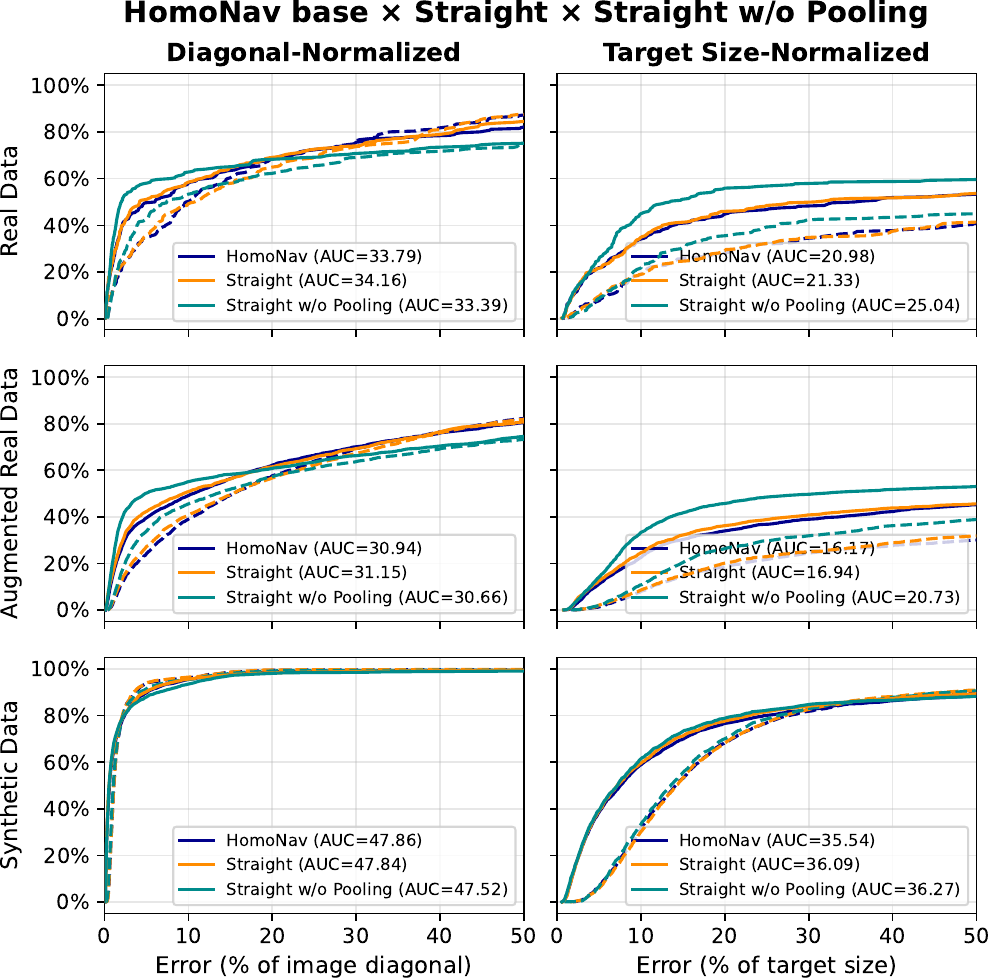}
    \caption{Comparison of architectural variants using cumulative error curves. Neck design ablation shows that the straight neck without pooling achieves the best performance at increased computational cost, while multi-resolution variants bring only marginal gains.}
    \label{fig:ablation-architecture}
\end{figure}

\myref{Table}{tab:results} and \myref{Figure}{fig:plot-baseline} compare HomoNav with SIFT+RANSAC and RoMa v2 \cite{edstedt2025romav2harderbetter} using cumulative error curves (ECDF), where higher and more left-shifted curves indicate better performance.
We report our most accurate variant: the straight-neck model without pooling, which produces higher-resolution outputs than \myref{Figure}{fig:cnn-architecture} at increased computational cost (see \myref{Section}{sec:ablation}).

A key difference is that SIFT and RoMa operate in a template-matching setting with known artifact identity and high-resolution inputs, whereas HomoNav jointly detects and localizes all artifacts from strongly downscaled images, trading precision for efficiency.
Dashed curves in \myref{Figure}{fig:plot-baseline} show pass~1 (global), while solid curves show pass~2 (refined; \myref{Section}{sec:two-pass-inference}). The second pass provides a significant improvement by re-sampling from the original resolution.

On real and augmented data, SIFT performs best, with RoMa and HomoNav moderately behind. On synthetic data, however, HomoNav clearly outperforms both baselines, indicating strong alignment with the training distribution and robustness to distortions and small object scales. This suggests better tolerance to scale variation, though more real, challenging data is needed to confirm this.



\subsection{Ablation: Stable Warp, Neighborhood, Neck Shape}
\label{sec:ablation}

\myref{Figure}{fig:ablation-training} and \myref{Figure}{fig:ablation-architecture} present cumulative error curves for several design choices. The first figure analyzes training strategies (Stable Warp and neighborhood supervision), while the second one compares different neck architectures.

In \myref{Figure}{fig:ablation-training}, removing Stable Warp (\myref{Section}{sec:two-pass-inference}) mainly degrades performance, most notably for pass~2, where this training is explicitly targeted. A smaller but visible drop is also observed for pass~1 on real data, likely because Stable Warp increases the number of near-canonical training examples matching the real capture conditions. Neighborhood supervision (\myref{Section}{sec:loss}) is beneficial for the multi-resolution neck, although its effect is moderate and dataset-dependent.

In \myref{Figure}{fig:ablation-architecture}, the straight neck without pooling gives the strongest target-size-normalized performance, especially on real data, while remaining comparable to the other variants in diagonal normalization. This gain comes at the cost of a higher-resolution output grid and increased computation. Interestingly, the straight neck with pooling performs nearly identically to the multi-resolution neck adapted from~\cite{tekin2018seamless}, suggesting that the additional high-resolution branch brings limited benefit in the current setup. This motivates further exploration of architectures that better trade off precision and efficiency.

\section{Conclusion}

We introduced homographic navigation, a geometry-driven framework that enables joint recognition and precise planar alignment from minimal supervision. Using synthetic homographic augmentation, we showed that accurate localization can be learned from a single reference image, while the proposed two-pass inference improves precision through coarse-to-fine refinement. A key contribution is the Stable Warp training strategy, which significantly improves performance -- especially for the high-precision second pass -- by aligning training conditions with the focused inference regime.

The present work addresses only the first phase of the full framework. Future work will focus on extending the approach by the second phase: by harvesting informative samples from unlabeled in-the-wild videos of artifacts and designing effective curriculum learning strategies. In parallel, we aim to develop lightweight models suitable for mobile deployment and to study the user experience of homographic navigation in augmented reality settings.

{
    \small
    \bibliographystyle{ieeenat_fullname}
    \bibliography{main}
}

\end{document}